\documentclass[10pt,twocolumn,letterpaper]{article}

\usepackage[accsupp]{axessibility}

\usepackage[pagenumbers]{cvpr} 

\usepackage{multirow}

\usepackage[dvipsnames]{xcolor}

\definecolor{cvprblue}{rgb}{0.21,0.49,0.74}
\usepackage[pagebackref,breaklinks,colorlinks,citecolor=cvprblue]{hyperref}

\def\paperID{1129} 
\def\confName{CVPR}
\def\confYear{2024}

\definecolor{tabblue}{HTML}{1f77b4}
\newcommand{\good}[1]{{\color{tabblue}{\textbf{#1}}}}
\newcommand{\bad}[1]{{\color{red}{\textbf{#1}}}}

\newcommand{\best}[1]{\textbf{{#1}}}

\newcommand{\raw}{{I}_{\text{R}}}
\newcommand{\comp}{{I}_{\text{C}}}
\newcommand{\enh}{{I}_{\text{E}}}
\newcommand{\ind}{{\perp\!\!\!\perp}}

\usepackage{tikz}
\usetikzlibrary{arrows.meta, shadows.blur, positioning}

\newlength{\vspaces}
\newlength{\vspacem}
\title{Enhancing Quality of Compressed Images by Mitigating Enhancement Bias Towards Compression Domain}

\author{
Qunliang Xing\textsuperscript{1},
Mai Xu\textsuperscript{1,}\thanks{Corresponding author.},
Shengxi Li\textsuperscript{1},
Xin Deng\textsuperscript{1},
Meisong Zheng,
Huaida Liu,
Ying Chen\\
\textsuperscript{1}Beihang University
}

\begin{document}
\maketitle

\begin{abstract}
    Existing quality enhancement methods for compressed images focus on aligning the enhancement domain with the raw domain to yield realistic images.
    However, these methods exhibit a pervasive enhancement bias towards the compression domain, inadvertently regarding it as more realistic than the raw domain.
    This bias makes enhanced images closely resemble their compressed counterparts, thus degrading their perceptual quality.
    In this paper, we propose a simple yet effective method to mitigate this bias and enhance the quality of compressed images.
    Our method employs a conditional discriminator with the compressed image as a key condition, and then incorporates a domain-divergence regularization to actively distance the enhancement domain from the compression domain.
    Through this dual strategy, our method enables the discrimination against the compression domain, and brings the enhancement domain closer to the raw domain.
    Comprehensive quality evaluations confirm the superiority of our method over other state-of-the-art methods without incurring inference overheads.
\end{abstract}

\section{Introduction}

With the emergence of the big data era, we have been witnessing an explosive growth in images and videos. 
According to statistics from Domo~\cite{inc_data_2020}, Instagram users shared approximately 66 thousand images every minute in 2022, marking an 18.3-fold increase over the past decade (\ie, 3.6 thousand in 2013).
Similar trends have been observed on other internet servers, including WeChat and Twitter.
To efficiently store and transmit this vast volume of images, several lossy image compression standards have been developed, such as joint photographic experts group (JPEG)~\cite{wallace_jpeg_1992}, JPEG 2000~\cite{marcellin_overview_2000}, and high-efficiency video coding (HEVC)~\cite{sullivan_overview_2012}.
However, images compressed with these standards inevitably suffer from compression artifacts, such as the effects of ringing, blocking, and blurring~\cite{shen_review_1998}.
These artifacts may significantly degrade the quality of user experience (QoE)~\cite{seshadrinathan_study_2010,itu-t_p10_2017}.

\begin{figure}
    \centering
    \includegraphics[width=\linewidth]{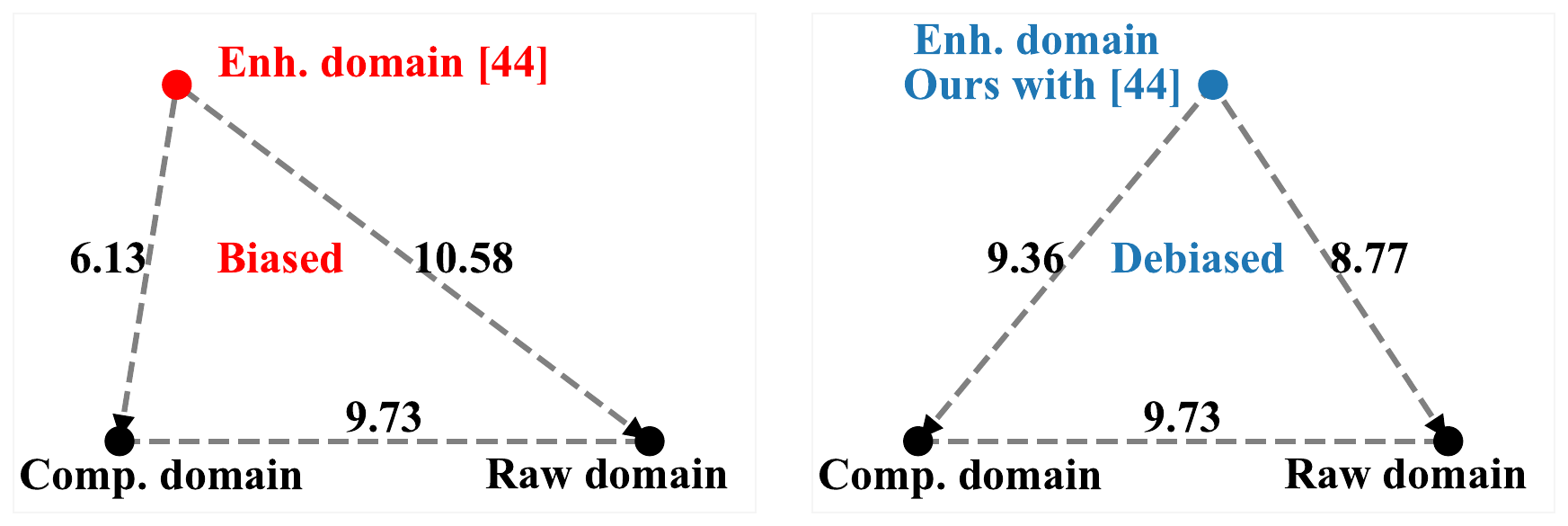}
    \includegraphics[width=\linewidth]{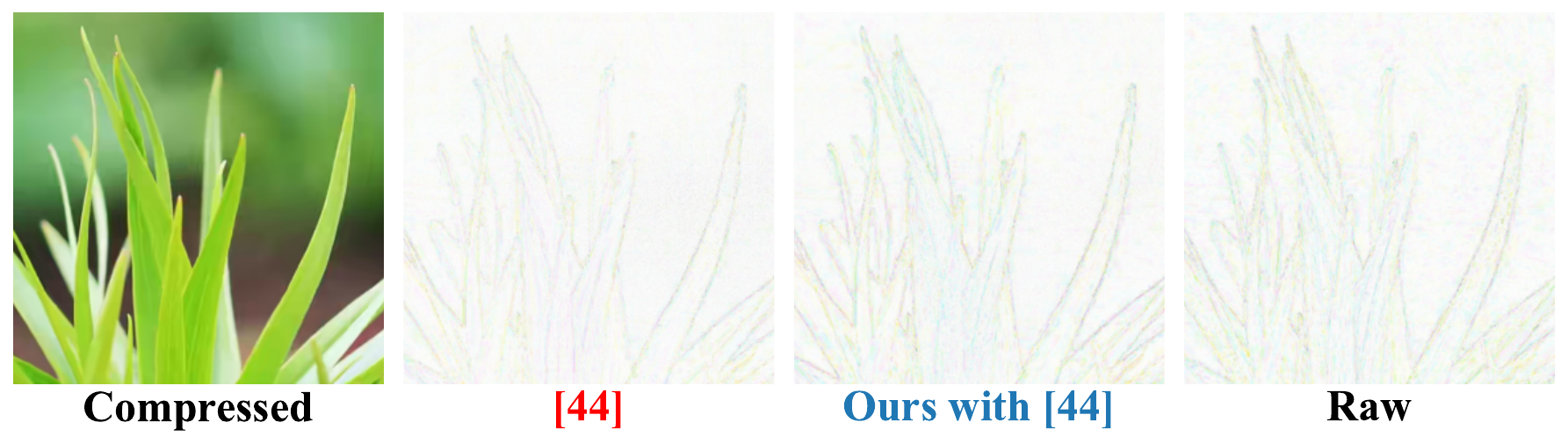}
    \caption{\textit{Top}: Frechet inception distance (FID)~\cite{heusel_gans_2017} scores between enhancement, compression, and raw domains on the DIV2K validation set~\cite{agustsson_ntire_2017}.
    \textit{Bottom}: Visualization of residual to the compressed image.
    The results illustrate the enhancement bias towards the compression domain.
    Our method effectively mitigates this bias, bringing the enhancement domain closer to the raw domain.\protect\footnotemark}
    \label{fig:fig1}
\end{figure}

\footnotetext{Three image domains correspond to the distributions of enhanced, raw, and compressed images.
Distances between domains are represented in triangle plots, adhering to the triangle inequality~\cite{wikipedia_contributors_triangle_2023} in our practice.}

To address the issue of quality degradation, a series of methods have been proposed to enhance the quality of compressed images.
Unfortunately, the majority of these methods~\cite{dong_compression_2015,wang_d3_2016,guo_building_2016,wang_novel_2017,zhang_learning_2017,xing_early_2020,xing_daqe_2023} focus on improving pixel-wise fidelity, limited in generating realistic textures as noted by Blau~\etal~\cite{blau_perception-distortion_2018}.
In response to this limitation, several methods~\cite{ghosh_iegan_2019,kim_towards_2020} have ventured into enhancing perceptual quality by predominantly leveraging perception-driven super-resolution (SR) baselines that employ generative adversarial networks (GANs)~\cite{goodfellow_generative_2014}.
By iteratively training a discriminator and a generator, the discriminator is optimized to discriminate between the enhancement and raw domains, and then guides the generator to produce realistic enhancement-domain images.

In fact, existing methods focus on aligning the enhancement domain with the raw domain, neglecting the crucial role of the compression domain in artifact reduction.
To delve into the ramification of this oversight, we conduct a comprehensive investigation into these methods.
Our findings highlight a pervasive enhancement bias towards the compression domain in two aspects.
1)
\textit{A deficiency in discriminating between the compression and raw domains}:
Despite the presence of compression artifacts, these methods perceive the compression domain as more realistic than the raw domain.
2)
\textit{A tendency to generate an enhancement domain more akin to the compression domain than to the raw domain}:
For existing methods, enhancement-domain images are more aligned with compression-domain images than with raw-domain images.
Given these insights, it becomes clear that the compression domain should not be overlooked;
instead, it is vital to harness its potential for enhancing image realism and quality.

Building upon these insights, we propose a simple yet effective method to mitigate the enhancement bias and enhance the quality of compressed images.
On one hand, our method discerns the compression domain by employing a conditional discriminator~\cite{mirza_conditional_2014} with the compressed image as a key condition.
We theoretically prove that the domain discrimination in GAN-based quality enhancement can be facilitated by our improved discriminator.
On the other hand, our method actively distinguishes the enhancement domain from the compression domain.
This distancing is achieved by a domain-divergence regularization that promotes the disparity between the enhancement and compression domains.
Through this dual strategy, our method enables the discrimination against the compression domain, and ultimately brings the enhancement domain closer to the raw domain to achieve superior image quality, as illustrated in \cref{fig:fig1}.
Finally, to verify the effectiveness of our proposed method, we conduct extensive experiments and affirm that our method proficiently addresses the problems highlighted in our observations, culminating in a substantial enhancement in image quality.

The main contributions of this paper are as follows.
\begin{enumerate}[label=\textbf{\arabic*)}]
    \item {We revisits the compression domain, an aspect often overlooked by existing methods.
    Through this exploration, we identify a significant enhancement bias towards the compression domain, which adversely affects the performance of quality enhancement.}
    \item {We introduce a new method aimed at mitigating this bias.
    This method, characterized by its simplicity and effectiveness, is model-agnostic and yields a marked improvement in image quality without incurring any inference overhead to existing methods.}
    \item {We conduct a series of comprehensive experiments including a thorough set of quality evaluations, demonstrating the effectiveness of the proposed method.}
\end{enumerate}

\section{Related Works}

In this section, we provide an overview of existing methods developed for enhancing the quality of compressed images.
Most of these methods~\cite{dong_compression_2015,wang_d3_2016,guo_building_2016,wang_novel_2017,zhang_learning_2017,xing_early_2020,xing_daqe_2023} have focused on improving compression fidelity using convolutional neural networks (CNNs).
For example, Dong~\etal~\cite{dong_compression_2015} introduced a pioneering CNN-based method for enhancing the quality of JPEG-compressed images through a shallow four-layer artifacts reduction-CNN (AR-CNN).
Following this, Zhang~\etal~\cite{zhang_learning_2017} developed the denoising CNN (DnCNN) that combines a 20-layer deep network with cutting-edge techniques such as residual learning~\cite{he_deep_2016} and batch normalization~\cite{ba_layer_2016}.
Wang~\etal~\cite{wang_novel_2017} designed a 10-layer deep CNN-based auto decoder (DCAD), signifying the first CNN-based quality enhancement method for HEVC-compressed images.
Then, Xing~\etal~\cite{xing_early_2020} proposed the resource-efficient blind quality enhancement (RBQE) method suitable for both JPEG and HEVC-compressed images, which uniquely incorporates a dynamic inference structure for blind yet effective quality enhancement.
Most recently, the defocus-aware quality enhancement (DAQE) method~\cite{xing_daqe_2023} emerged to initially discern region-wise quality differences via defocus estimation, followed by a region-wise divide-and-conquer enhancement.
However, these fidelity-oriented methods, despite their advancements, still face challenges in generating realistic textures and often produce over-smoothed images~\cite{blau_perception-distortion_2018}.

To address the challenges of fidelity-oriented methods, Ghosh~\etal~\cite{ghosh_iegan_2019} and Kim~\etal~\cite{kim_towards_2020} have focused on enhancing the perceptual quality of compressed images.
They both leverage super-resolution (SR) methods~\cite{ledig_photo-realistic_2017,wang_esrgan_2018} as baselines by incorporating different GANs~\cite{goodfellow_generative_2014}.
Specifically, Ghosh~\etal~\cite{ghosh_iegan_2019} proposed the image enhancement GAN (IEGAN), a method grounded in the SRGAN method~\cite{ledig_photo-realistic_2017} with a standard GAN, for enhancing the quality of JPEG-compressed images.
In \cite{kim_towards_2020}, Kim~\etal leveraged the enhanced SRGAN (ESRGAN) method~\cite{wang_esrgan_2018} for quality enhancement by inheriting the relativistic average GAN (RaGAN)~\cite{jolicoeur-martineau_relativistic_2019} and all loss functions.
Recently, the Real-ESRGAN method~\cite{wang_real-esrgan_2021} with a vanilla GAN regularized by the spectral normalization~\cite{miyato_spectral_2018} has demonstrated its capability in enhancing real-world images that contain compression artifacts.

Unfortunately, the aforementioned quality enhancement methods neglect the compression domain for reducing artifacts, since they focus intently on aligning the enhancement domain with the raw domain, in a manner akin to the prevalent SR methods.
Consequently, their enhanced images exhibit a strong resemblance to the compressed counterparts, leading to an enhancement bias towards the compression domain. 
We shall delve into this bias in the subsequent section.

\section{Enhancement Bias Towards Compression Domain}

\begin{table}
    \centering
    \small
    \begin{tabular}{@{}ll|c|cc|cc@{}}
        \toprule
        \multirow{2}{*}{{Meth.}} & \multirow{2}{*}{{Data.}} & \multirow{2}{*}{{Raw}} & \multirow{2}{*}{{Enh.}} & {$\Delta$} & \multirow{2}{*}{{Comp.}} & {$\Delta$} \\
        & & & & {to raw} & & {to raw} \\
        \midrule
        \multirow{2}{*}{\cite{wang_real-esrgan_2021}} & DIV2K & 0.76 & 0.28 & \good{-0.48} & 0.84 & \bad{+0.08} \\
        & Flickr2K & 0.74 & 0.27 & \good{-0.47} & 0.84 & \bad{+0.10} \\
        \midrule
        \multirow{2}{*}{\cite{wang_esrgan_2018}} & DIV2K & 0.73 & 0.27 & \good{-0.46} & 0.76 & \bad{+0.03} \\
        & Flickr2K & 0.74 & 0.26 & \good{-0.48} & 0.78 & \bad{+0.04} \\
        \bottomrule
    \end{tabular}
    \caption{Discriminator-evaluated realism scores for raw, enhanced, and compressed images.
    Higher scores indicate greater perceived realism.
    The results reveal that existing methods regard compressed images as more realistic than raw images.}
    \label{tab:observation:discrim}
    \vspace{\vspacem}
\end{table}

In this section, we thoroughly examine the enhancement bias towards the compression domain, a phenomenon we consistently observe across existing perception-driven quality enhancement methods for compressed images.
As aforementioned, these methods predominantly leverage perception-driven SR methods as baselines.
We thus present the analysis results based on \cite{wang_esrgan_2018,wang_real-esrgan_2021}, given their prominence in recent perception-driven quality enhancement endeavors.
Note that these methods are substantially different:
The former method~\cite{wang_esrgan_2018} utilizes a VGG~\cite{simonyan_very_2015}-based RaGAN~\cite{jolicoeur-martineau_relativistic_2019}, while the latter method~\cite{wang_real-esrgan_2021} adopts a U-Net~\cite{schonfeld_u-net_2020}-based vanilla GAN~\cite{goodfellow_generative_2014} with spectral normalization regularization~\cite{miyato_spectral_2018}.
We re-train these methods by their default settings for quality enhancement using the images from the DIV2K training set~\cite{agustsson_ntire_2017}.
Subsequently, these methods are evaluated using the images from the DIV2K validation and Flickr2K~\cite{timofte_ntire_2017} datasets.
All images are compressed by better portable graphics (BPG)~\cite{bellard_better_2018} and JPEG~\cite{wallace_jpeg_1992}.\footnote{BPG implements the HEVC intra-frame compression~\cite{sullivan_overview_2012}.
Results for BPG-compressed images with quality parameter (QP) set to 37 are presented, while other results are comprehensively provided in the supplementary material.}

\textbf{Observation 1:}
Existing quality enhancement methods fail to discriminate against compressed images, regarding them as more realistic than raw images.

\begin{figure}
    \centering
    \includegraphics[width=.95\linewidth]{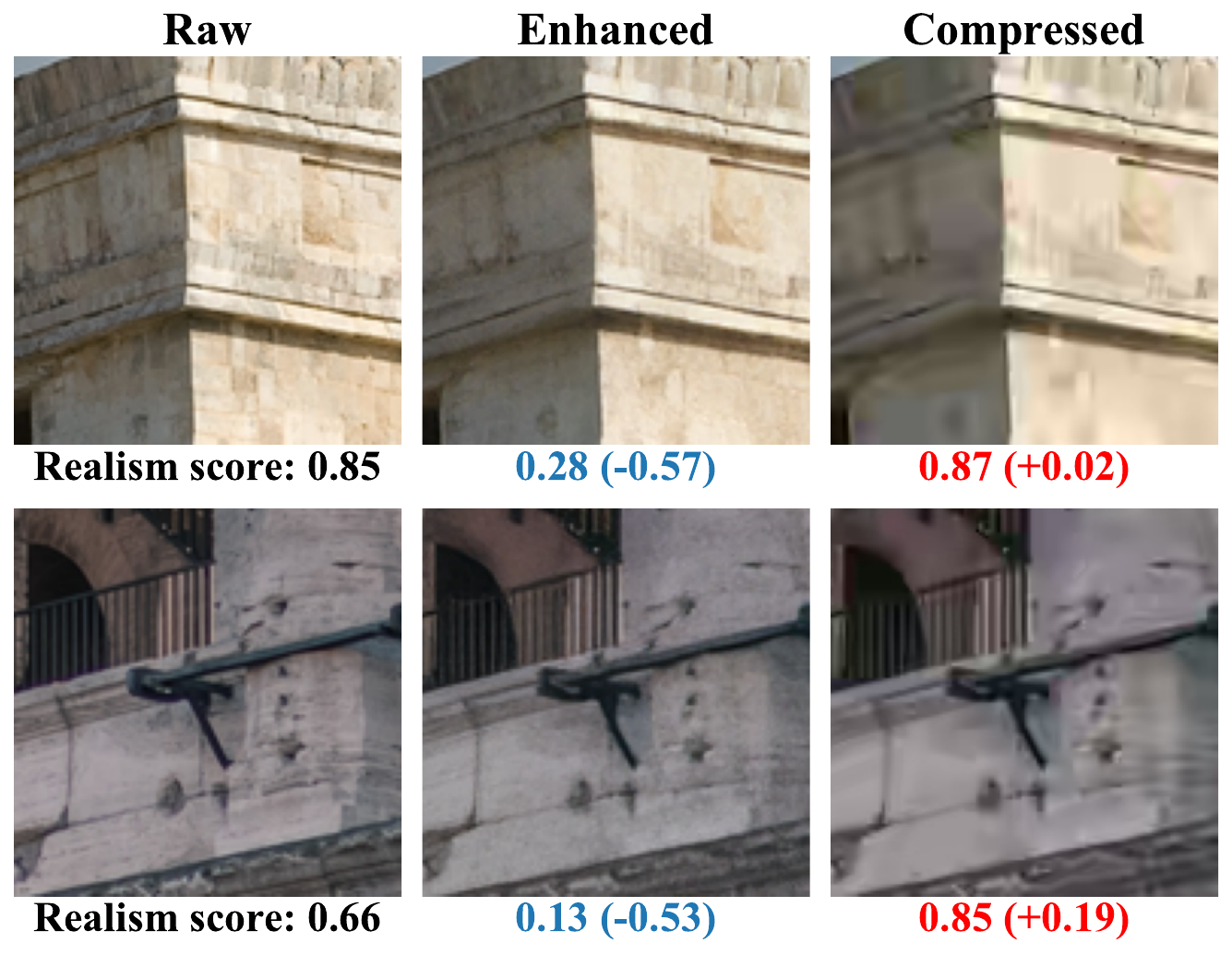}
    \caption{Visual demonstrations of realism scores for images as evaluated by \cite{wang_real-esrgan_2021}.
    The results highlight that existing methods overlook significant compression artifacts in realism evaluation.}
    \label{fig:observation:discrim_demo}
    \vspace{\vspacem}
\end{figure}

\textit{Analysis.}
We measure the realism scores of enhanced, compressed, and raw images evaluated by discriminators.
Specifically, we crop all images into patches with the size of $128 \times 128$ (in accordance with the patch size during training), which are then sent to the discriminators for evaluating realism scores.
As shown in \cref{tab:observation:discrim}, the discriminators effectively discriminate between enhanced and raw images by assigning higher realism scores to the raw images.
However, these discriminators struggle to distinguish between compressed and raw images;
more surprisingly, the average realism score for compressed images even exceeds that of raw images.
This reveals that the compression domain is perceived as more realistic, undermining the efforts to reduce compression artifacts and leading to the enhancement bias towards the compression domain.
\cref{fig:observation:discrim_demo} provides visual demonstrations to further illustrate our observation that the severe compression artifacts (\eg, blurring) are neglected when evaluating image realism.
We argue that these discriminators predominantly base their judgment on generative artifacts;
as a result, the compressed image appears more realistic since the introduced compression artifacts differ significantly from generative artifacts.
This insight wraps up our analysis for Observation 1.

\textbf{Observation 2:}
For existing methods, enhancement-domain images are more aligned with compression-domain images than with raw-domain images.

\begin{figure*}
    \centering
    \includegraphics[width=\linewidth]{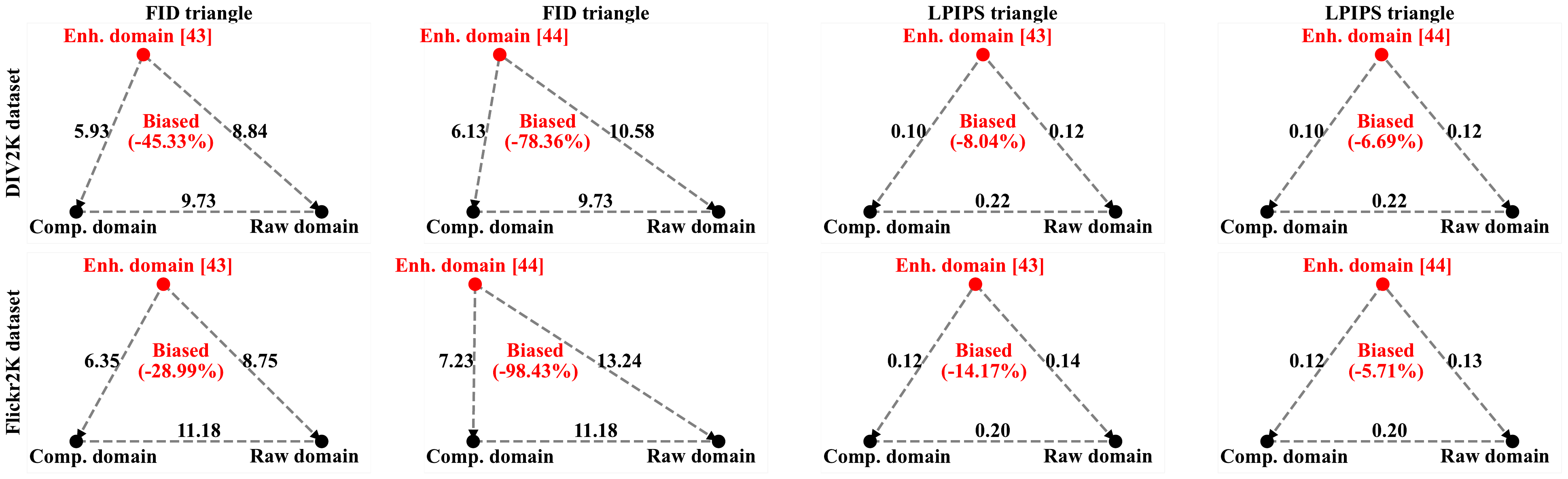}
    \caption{Similarity scores between compression, raw, and enhancement domains.
    Lower FID and LPIPS scores indicate greater similarity.
    The horizontal deviation of each vertex relative to the centroid of its base, calculated with float precision on original data, is also shown.
    The results underscore the enhancement bias, demonstrating a closer alignment of the enhancement domain with the compression domain than with the raw domain.}
    \label{fig:observation:triangles}
    \vspace{\vspacem}
\end{figure*}

\begin{figure}
    \centering
    \includegraphics[width=.95\linewidth]{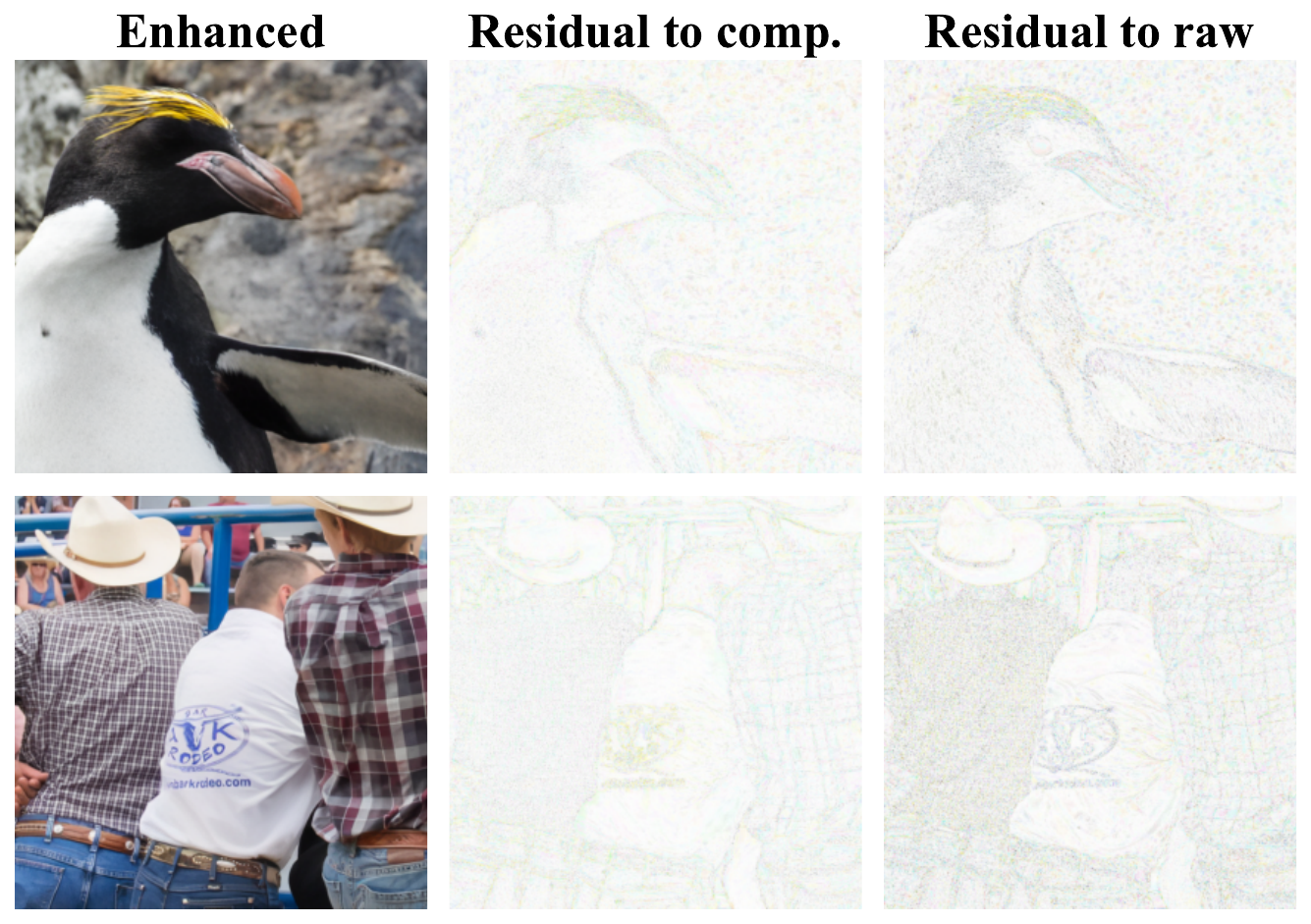}
    \caption{Residual comparisons between enhanced images~\cite{wang_real-esrgan_2021} and their compressed/raw counterparts.
    These visualizations reveal a stronger resemblance of the enhanced images to their compressed counterparts, as indicated by the weaker residual.}
    \label{fig:observation:residual}
    \vspace{\vspacem}
\end{figure}

\textit{Analysis.}
We analyze the similarity scores among compressed, raw, and enhanced images via utilizing the datasets and re-trained models above.
Here, the similarity is measured by two widely-adopted metrics for assessing perceptual quality enhancement: the Frechet inception distance (FID)~\cite{heusel_gans_2017} and the learned perceptual image patch similarity (LPIPS)~\cite{zhang_unreasonable_2018}.
As illustrated in \cref{fig:observation:triangles}, there is a consistent trend where the enhancement domain exhibits closer alignment with the compression domain, compared to the raw domain.
To quantitatively measure this trend, we calculate the horizontal deviation of each vertex relative to the centroid of its base, which can be equivalently expressed as:
\begin{equation}
    \text{Deviation} := \frac{S_{\text{C}, \text{E}}^2 - S_{\text{R}, \text{E}}^2}{S_{\text{C}, \text{R}}^2} \times 100\%,
\end{equation}
where $S_{\text{C}, \text{E}}$, $S_{\text{C}, \text{R}}$ and $S_{\text{R}, \text{E}}$ represent the similarity (in terms of either FID or LPIPS) between compression and enhancement domains, between compression and raw domains, and between raw and enhancement domains, respectively.
The obtained results range from -5.71\% to -98.43\%, demonstrating the pervasive bias towards the compression domain, as observed consistently across datasets and similarity metrics.
\cref{fig:observation:residual} provides visual demonstrations to further illustrate our observation.
This comprehensive analysis brings us to the conclusion of Observation 2.

\section{Mitigating Enhancement Bias for Quality Enhancement of Compressed Images}

Given the above observations, we propose a simple yet effective method that aims at mitigating the enhancement bias to improve the quality of compressed images.
First, our method discerns the compression domain by utilizing a conditional discriminator~\cite{mirza_conditional_2014}, wherein the compressed image serves as a pivotal condition.
Second, our method actively distinguishes the enhancement domain from the compression domain via a proposed domain-divergence regularization.
Employing this dual strategy, our method facilitates effective discrimination against the compression domain, and consequently aligns the enhancement domain more closely with the raw domain.
Ultimately, by mitigating the enhancement bias, our method significantly enhances the quality of compressed images.

\subsection{Discrimination against Compression Domain}

Existing methods predominantly utilize GANs~\cite{goodfellow_generative_2014} to synthesize realistic images.
Specifically, given a compressed image $\comp$, these methods focus on training a generative network $G(\cdot)$, which produces an enhanced image denoted as $\enh = G(\comp)$.
Simultaneously, an image-realism discriminator $D(\cdot)$, designed to distinguish between the enhanced image $\enh$ and the raw image $\raw$, is optimized via an adversarial loss function:
\begin{equation}
    \max_D \mathcal{L}_{D} = \mathbb{E}[ \log \big( 1 - D {( \enh )} \big)] +  \mathbb{E}[\log D {( \raw )}].
\end{equation}
Here, the discriminator aims to output $1$ for $\raw$ and $0$ for $\enh$.

However, we note that $\comp$ is derived from compressing $\raw$ and is subsequently used to produce $\enh$, suggesting that $\raw$ and $\enh$ have correlated distributions.
Contrarily, most GANs are established on the principle of independent generated and real image distributions, typically with generated images sampled from random noise~\cite{goodfellow_generative_2014}.
The violation of this independence may lead to mode collapse~\cite{li_iid-gan_2023,arjovsky_towards_2017}, resulting in enhanced images that are deficient in realism.
This outcome aligns with what we have identified as biased enhancement in our observations.

\begin{figure}
    \centering
    \begin{tikzpicture}[node distance=1.5cm,
        node/.style={circle, fill=white, blur shadow={shadow blur steps=5}, draw=black, text centered, thick, minimum size=0.75cm},
        emphasized node/.style={circle, fill=white, blur shadow={shadow blur steps=5}, draw=red, text=red, text centered, thick, minimum size=0.75cm},
        every path/.style={->, >=Stealth, thick}
    ]
        \node[node] (X) {$\raw$};
        \node[emphasized node, right=of X] (Y) {$\comp$};
        \node[node, right=of Y] (Z) {$\enh$};
        
        \draw (X) -- (Y) node[midway, above] {$f_c$};
        \draw (Y) -- (Z) node[midway, above] {$f_e$};
    \end{tikzpicture}
    \caption{Causal relationships among the raw, compressed, and enhanced images, where $f_c$ and $f_e$ denote the image compression and quality enhancement respectively.}
    \label{fig:method:relationship}
    \vspace{\vspaces}
\end{figure}
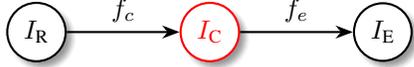

Indeed, $\raw$, $\comp$, and $\enh$ constitute a basic probabilistic graphical model (PGM), comprising a three-node directed acyclic graph (DAG)~\cite{thulasiraman_graphs_1992} as depicted in \cref{fig:method:relationship}. 
Recall that $\comp$ results from compressing $\raw$ and is later enhanced to form $\enh$.
This underlying causal graph underpins the Bayesian D-separation relationship:
\begin{equation}\label{eq:d_sep}
    p(\raw, \enh | \comp) \!=\! \underbrace{p(\raw | \comp) \!\cdot\! p(\enh | \raw, \comp)}_{\text{Chain rule}} \!=\! \underbrace{p(\raw | \comp) \!\cdot\! p(\enh | \comp)}_{\text{Causality}},
\end{equation}
thereby indicating $\raw$ and $\enh$ are conditionally independent given $\comp$, denoted as $\raw \ind \enh \mid \comp$.

\begin{table}
    \centering
    \small
    \begin{tabular}{@{}l|c|cc|cc@{}}
        \toprule
        \multirow{2}{*}{{Data.}} & \multirow{2}{*}{{Raw}} & \multirow{2}{*}{{Enh.}} & {$\Delta$} & \multirow{2}{*}{{Comp.}} & {$\Delta$} \\
        & & & {to raw} & & {to raw} \\
        \midrule
        DIV2K & 0.87 & 0.04 & \good{-0.83} & 0.05 & \good{-0.82} \\
        Flickr2K & 0.86 & 0.03 & \good{-0.83} & 0.05 & \good{-0.81} \\
        \bottomrule
    \end{tabular}
    \caption{Realism scores evaluated by our method using \cite{wang_real-esrgan_2021} for raw, enhanced, and compressed images.
    Higher scores indicate greater perceived realism.}
    \label{tab:method:discrim}
    \vspace{\vspacem}
\end{table}

This insight allows us to treat the enhanced and raw images as independent variables when $\comp$ is provided for training both enhancement and discrimination networks.
Thus, we deploy a conditional discriminator~\cite{mirza_conditional_2014}, incorporating the compression image as a key condition for quality enhancement:
\begin{equation}
    \max_D \mathcal{L}_{D} = \mathbb{E}[\log (1 - D {(\enh | \comp )}) ] + \mathbb{E}[\log D {( \raw | \comp )}].
\end{equation}
Empirical evidence strongly supports our method's effectiveness;
providing the discriminator with concatenated inputs of compressed and enhanced/raw images markedly enhances realism evaluation, as evidenced in \cref{tab:method:discrim}. 
More importantly, the enhanced discriminator effectively distinguishes the compression domain, which is achieved without modifying network architectures or optimization strategies.

\subsection{Distancing from Compression Domain}

Existing methods leverage a weighted combination of three distinct loss functions to optimize the generator $G(\cdot)$:
\begin{equation}
    \mathcal{L}_{G} = \lambda_{\text{r}} \mathcal{L}_{\text{recon.}} + \lambda_{\text{p}} \mathcal{L}_{\text{percept.}} + \lambda_{\text{d}} \mathcal{L}_{\text{discrim.}}.
\end{equation}
In this formulation, $\mathcal{L}_{\text{recon.}}$ and $\mathcal{L}_{\text{percept.}}$ assess the pixel-wise and feature-wise mean absolute error (MAE) respectively, both between $\enh$ and $\raw$;
$\mathcal{L}_{\text{discrim.}}$ quantifies the discrimination loss associated with $\enh$.
Notably, the compressed image $\comp$, indicative of the presence of compression artifacts, is largely overlooked.
However, as outlined in Observation 2, enhancement-domain images in these methods are more aligned with compression-domain images than with raw-domain images.

\begin{figure}
    \centering
    \includegraphics[width=\linewidth]{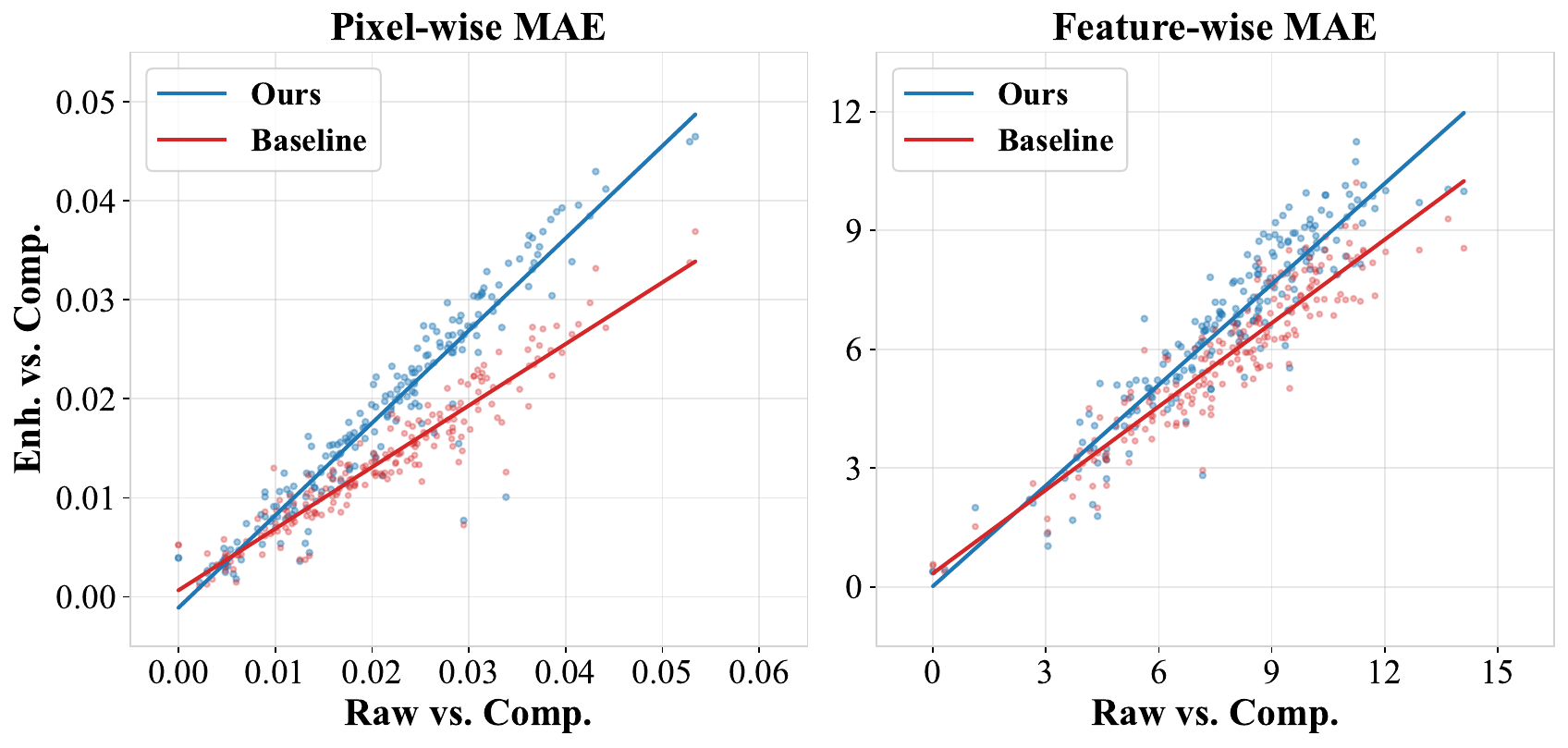}
    \caption{Visualization of domain disparity by our method using \cite{wang_real-esrgan_2021}, as applied to randomly cropped patches from the DIV2K validation set.
    Linear regression lines are also illustrated.}
    \label{fig:method:diff}
    \vspace{\vspacem}
\end{figure}

To address this challenge, we incorporate a domain-divergence regularization that accounts for the disparities from the compression domain to both the raw and enhancement domains:
\begin{gather}
    \mathcal{L}_R = 
    \begin{cases} 
        {\mathcal{D}}_{\text{C}, \text{R}} - {\mathcal{D}}_{\text{C}, \text{E}}, &\text{if } {\mathcal{D}}_{\text{C}, \text{E}} < {\mathcal{D}}_{\text{C}, \text{R}} \\
        0, &\text{otherwise} 
    \end{cases},\\
    {\mathcal{D}}_{\text{C}, \text{R}} = {| \psi_l (\raw) - \psi_l (\comp) |},\\
    {\mathcal{D}}_{\text{C}, \text{E}} = {| \psi_l (\enh) - \psi_l (\comp) |}.
\end{gather}
Here, $\psi_l$ represents the final convolution layer within the $l$-th block of a pre-trained VGG-19 model~\cite{simonyan_very_2015}.
During the training process, this regularization promotes a larger disparity between the enhancement and compression domains, especially when it is less pronounced than the disparity between the raw and compression domains;
otherwise, this regularization exerts no influence on training.
Consequently, the loss function evolves to:
\begin{equation}
    {\mathcal{L}}_{G} = \lambda_{\text{r}} \mathcal{L}_{\text{recon.}} + \lambda_{\text{p}} \mathcal{L}_{\text{percept.}} + \lambda_{\text{d}} \mathcal{L}_{\text{discrim.}} + \lambda_{\mathcal{R}} \mathcal{L}_R.
\end{equation}
As substantiated in \cref{fig:method:diff}, this regularization effectively enlarges the disparity between the compression and enhancement domains, both pixel-wise and feature-wise, to a degree comparable with the compression-to-raw domain disparity.

\textbf{Discussion.}
While many conditional-discriminator-based methods have shown advancements in tasks such as underwater enhancement~\cite{yu_underwater-gan_2019}, de-raining~\cite{zhang_image_2020}, interactive editing~\cite{cai_toward_2021}, and neural codecs~\cite{mentzer_high-fidelity_2020,yang_perceptual_2022}, this paper pioneers advances in the field of quality enhancement.
It also provides a unique perspective to understand these advancements through our bias analysis.
Our proposed method proficiently addresses the challenges of enhancement bias in a simple yet highly effective manner.
Its model-agnostic nature facilitates seamless integration with a wide range of enhancement methods, without incurring inference overheads.
Furthermore, an extensive array of SR methods can be enhanced and employed as baselines for perception-driven quality enhancement.
We further investigate the efficacy of our method for quality enhancement in the subsequent section.

\begin{table*}
    \centering
    \small
    \begin{tabular}{@{}l|cccccccccccc@{}}
		\toprule
		\multirow{2}{*}{Metric} & \multirow{2}{*}{{Comp.}} & \multirow{2}{*}{{\cite{dong_compression_2015}}} & \multirow{2}{*}{{\cite{wang_novel_2017}}} & \multirow{2}{*}{{\cite{zhang_learning_2017}}} & \multirow{2}{*}{{\cite{guo_toward_2019}}} & \multirow{2}{*}{{\cite{zhang_residual_2018}}} & \multirow{2}{*}{{\cite{xing_early_2020}}} & \multirow{2}{*}{{\cite{zamir_multi-stage_2021}}} & \multirow{2}{*}{{\cite{wang_esrgan_2018}}} & Ours & \multirow{2}{*}{{\cite{wang_real-esrgan_2021}}} & Ours \\
		 & & & & & & & & & & \scriptsize{w/ {\cite{wang_esrgan_2018}}} & & \scriptsize{w/ {\cite{wang_real-esrgan_2021}}} \\
		\midrule
		$\uparrow$ AHIQ~\cite{lao_attentions_2022} & .433 & .443 & .451 & .453 & .459 & .450 & .454 & .454 & .456 & .463 & .454 & \best{.471} \\
		$\uparrow$ CLIP.~\cite{wang_exploring_2023} & .494 & .538 & .548 & .553 & .572 & .551 & .556 & .558 & .662 & \best{.672} & .645 & .660 \\
		$\downarrow$ DISTS~\cite{ding_image_2020} & .126 & .134 & .143 & .143 & .140 & .144 & .142 & .141 & .060 & \best{.058} & .064 & .062 \\
		$\downarrow$ FID~\cite{heusel_gans_2017} & 9.73 & 10.3 & 10.4 & 10.7 & 10.7 & 10.6 & 10.3 & 10.3 & 8.84 & \best{7.50} & 10.6 & 8.77 \\
		$\uparrow$ Hyper.~\cite{su_blindly_2020} & .459 & .507 & .541 & .549 & .560 & .544 & .545 & .553 & .557 & .576 & .563 & \best{.580} \\
		$\downarrow$ LPIPS~\cite{zhang_unreasonable_2018} & .218 & .215 & .214 & .212 & .206 & .213 & .210 & .208 & .120 & .114 & .116 & \best{.113} \\
		$\uparrow$ MUSIQ~\cite{ke_musiq_2021} & 60.7 & 62.2 & 63.0 & 63.1 & 63.7 & 63.0 & 63.1 & 63.3 & 65.1 & 65.6 & 64.9 & \best{66.2} \\
		$\downarrow$ NIQE~\cite{mittal_making_2013} & 4.22 & 4.49 & 4.61 & 4.60 & 4.63 & 4.65 & 4.61 & 4.65 & 2.87 & \best{2.80} & 3.14 & 2.92 \\
		$\downarrow$ PI~\cite{blau_2018_2018} & 4.35 & 4.52 & 4.59 & 4.58 & 4.60 & 4.61 & 4.58 & 4.60 & 3.15 & \best{3.12} & 3.43 & 3.24 \\
		$\uparrow$ PSNR (dB) & 30.8 & 31.2 & 31.4 & 31.5 & \best{31.7} & 31.5 & 31.5 & 31.6 & 29.0 & 29.6 & 29.8 & 30.6 \\
		$\uparrow$ TOPIQ~\cite{chen_topiq_2023} & .721 & .716 & .713 & .713 & .724 & .711 & .716 & .720 & .811 & \best{.822} & .810 & .821 \\
		\bottomrule
    \end{tabular}
    \caption{Objective quality of enhanced BPG-compressed images with QP set to 37.
    Results for other QPs and JPEG-compressed images, following similar trends, are detailed in the supplementary material.
    All results are to three significant figures.}
    \label{tab:exp:objective_quality}
    \vspace{\vspacem}
\end{table*}

\section{Experiments}

In this section, we present experimental results to verify the performance of our proposed method for the quality enhancement of compressed images.

\subsection{Settings}

\textbf{Image dataset and compression.}
We utilize the high-quality raw images from the DIV2K dataset~\cite{agustsson_ntire_2017} for training and evaluation.
Given the widespread adoption of BPG~\cite{bellard_better_2018} and JPEG~\cite{wallace_jpeg_1992} as image compression codecs, we compress all images using these two codecs.
Specifically, we employ five different compression settings for each codec: the QP for BPG is set to 27, 32, 37, 42, and 47, whereas the quality factor (QF) for JPEG is set to 10, 20, 30, 40, and 50.
These settings are in alignment with the common practices in previous quality enhancement works~\cite{yang_multi-frame_2018,guan_mfqe_2021,zheng_progressive_2022}.

\textbf{Baselines and metrics.}
Our method is compared with a range of widely-used quality enhancement methods\footnote{\url{https://github.com/ryanxingql/powerqe}}, including AR-CNN~\cite{dong_compression_2015}, DCAD~\cite{wang_novel_2017}, DnCNN~\cite{zhang_learning_2017}, CBDNet~\cite{guo_toward_2019}, RDN~\cite{zhang_residual_2018}, RBQE~\cite{xing_early_2020}, and MPRNet~\cite{zamir_multi-stage_2021}.
All of these methods are re-trained using our dataset for a fair comparison.
In addition, GAN-based SR methods~\cite{wang_esrgan_2018,wang_real-esrgan_2021}\footnote{\url{https://github.com/XPixelGroup/BasicSR}} are adopted as baselines given their prominence in recent perception-driven quality enhancement endeavors.
Accordingly, our training settings are consistent with these methods.
Specifically, $\lambda_{\text{r}}$, $\lambda_{\text{p}}$, $\lambda_{\text{d}}$, and $\lambda_{\mathcal{R}}$ are set to $1e{-2}$, $1$, $5e{-3}$, and $1e{-1}$ respectively for \cite{wang_esrgan_2018}, and are set to $1$, $1$, $1e{-1}$, and $1e{-1}$ respectively for \cite{wang_real-esrgan_2021}.
The evaluation is conducted using a comprehensive set of metrics\footnote{\url{https://github.com/chaofengc/IQA-PyTorch}}, namely AHIQ~\cite{lao_attentions_2022}, CLIP-IQA~\cite{wang_exploring_2023}, DISTS~\cite{ding_image_2020}, FID~\cite{heusel_gans_2017}, HyperIQA~\cite{su_blindly_2020}, LPIPS~\cite{zhang_unreasonable_2018}, MUSIQ~\cite{ke_musiq_2021}, NIQE~\cite{mittal_making_2013}, PI~\cite{blau_2018_2018}, PSNR, and TOPIQ~\cite{chen_topiq_2023}.

\subsection{Evaluation}

\begin{figure}
    \centering
    \includegraphics[width=\linewidth]{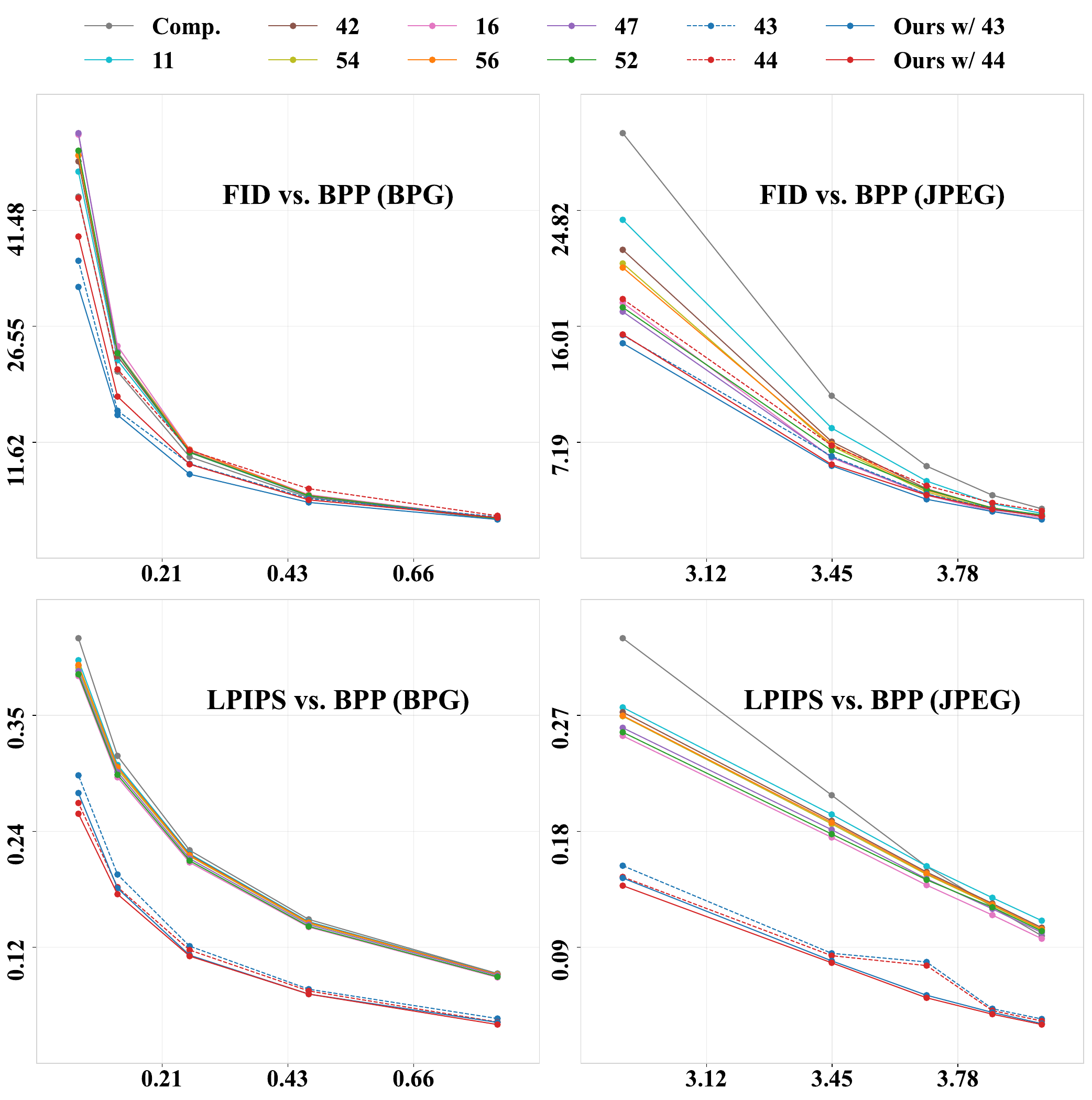}
    \caption{Rate-distortion curves comparing bits per pixel (BPP) against distortion measured by FID and LPIPS.
    Additional results using other metrics are provided in the supplementary material.}
    \label{fig:exp:rd_curve}
    \vspace{\vspacem}
\end{figure}

\begin{table*}
    \centering
    \small
    \begin{tabular}{@{}l|ccccccccccc@{}}
		\toprule
		\multirow{2}{*}{Metric} & \multirow{2}{*}{{\cite{dong_compression_2015}}} & \multirow{2}{*}{{\cite{wang_novel_2017}}} & \multirow{2}{*}{{\cite{zhang_learning_2017}}} & \multirow{2}{*}{{\cite{guo_toward_2019}}} & \multirow{2}{*}{{\cite{zhang_residual_2018}}} & \multirow{2}{*}{{\cite{xing_early_2020}}} & \multirow{2}{*}{{\cite{zamir_multi-stage_2021}}} & \multirow{2}{*}{{\cite{wang_esrgan_2018}}} & Ours & \multirow{2}{*}{{\cite{wang_real-esrgan_2021}}} & Ours \\
		 & & & & & & & & & \scriptsize{w/ {\cite{wang_esrgan_2018}}} & & \scriptsize{w/ {\cite{wang_real-esrgan_2021}}} \\
		\midrule
		AHIQ~\cite{lao_attentions_2022} & -5.63 & -10.4 & -12.0 & -18.8 & -11.6 & -15.6 & -15.6 & -24.6 & -29.3 & -28.3 & \best{-41.5} \\
		CLIP.~\cite{wang_exploring_2023} & -29.5 & -38.1 & -40.8 & -55.7 & -42.9 & -46.0 & -47.2 & -93.2 & -93.5 & -94.8 & \best{-96.8} \\
		DISTS~\cite{ding_image_2020} & +12.4 & +26.5 & +26.4 & +21.6 & +28.9 & +26.2 & +23.9 & -75.5 & \best{-75.8} & -69.5 & -72.0 \\
		FID~\cite{heusel_gans_2017} & +6.73 & +9.01 & +11.5 & +14.9 & +11.0 & +11.8 & +10.7 & -21.7 & \best{-24.3} & +3.64 & -13.7 \\
		Hyper.~\cite{su_blindly_2020} & -46.3 & -62.9 & -65.9 & -74.6 & -65.8 & -66.6 & -68.9 & -82.1 & \best{-98.5} & -83.5 & -95.3 \\
		LPIPS~\cite{zhang_unreasonable_2018} & -5.44 & -6.64 & -7.92 & -12.6 & -6.90 & -9.66 & -11.4 & -61.2 & -63.5 & -64.7 & \best{-66.2} \\
		MUSIQ~\cite{ke_musiq_2021} & -24.7 & -34.1 & -35.0 & -42.5 & -35.2 & -36.3 & -38.9 & -71.9 & \best{-81.6} & -64.2 & -77.9 \\
		PSNR & -11.3 & -15.3 & -16.7 & \best{-20.6} & -16.3 & -18.3 & -19.0 & +58.7 & +36.9 & +30.7 & +5.71 \\
		TOPIQ~\cite{chen_topiq_2023} & -1.46 & -1.65 & -2.03 & -7.05 & -1.48 & -3.55 & -5.19 & -36.0 & -39.8 & -36.7 & \best{-40.3} \\
		\bottomrule
    \end{tabular}
    \caption{BD-BR (\%) performance applied to BPG-compressed images.
    Negative values indicate a reduction in bit rate for the same quality.
    Results for JPEG-compressed images are available in the supplementary material.
    All results are to three significant figures.}
    \label{tab:exp:bdbr}
    \vspace{\vspaces}
\end{table*}

\begin{figure*}
    \centering
    \includegraphics[width=\linewidth]{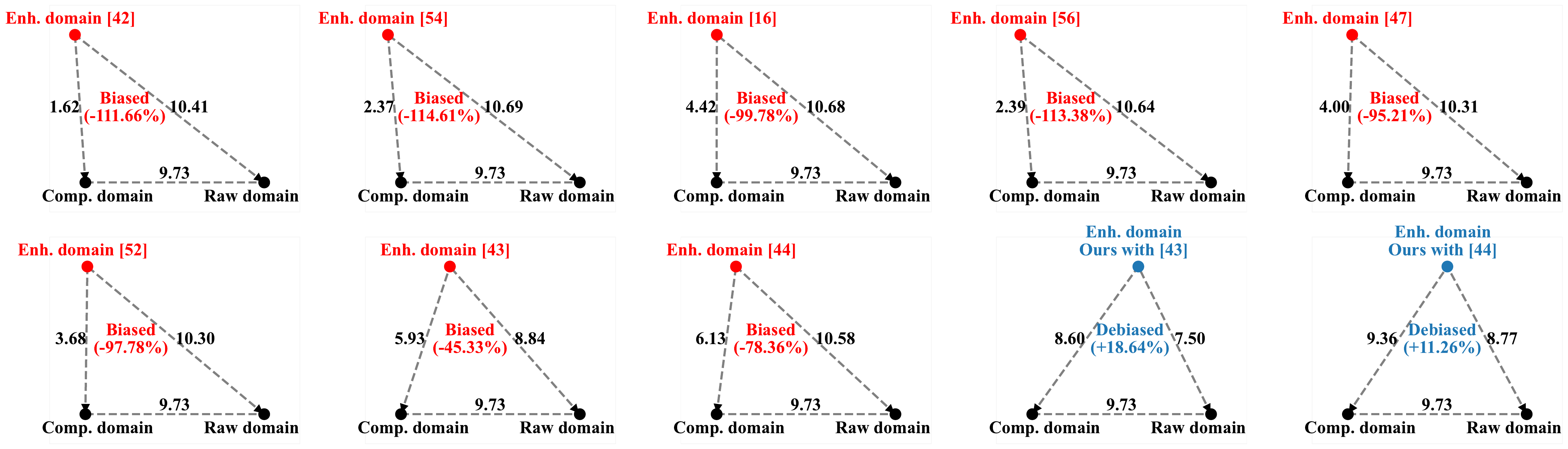}
    \caption{FID scores between three domains of the evaluation set compressed by BPG with QP set to 37.
    The horizontal deviation of each vertex relative to the centroid of its base, calculated with float precision on original data, is also shown.}
    \label{fig:exp:bias_qp37_fid}
    \vspace{\vspacem}
\end{figure*}

\textbf{Objective quality evaluation.}
\cref{tab:exp:objective_quality} presents the objective quality of enhanced compressed images.
As detailed in \cref{tab:exp:objective_quality}, our method consistently enhances the perception-driven SR baseline methods~\cite{wang_esrgan_2018,wang_real-esrgan_2021} across all metrics.
For instance, the FID score by \cite{wang_real-esrgan_2021} stands at 10.6, which is even higher than the 9.73 baseline for compressed images, indicating a post-enhancement degradation in fidelity to raw images.
On contrary, our proposed method effectively addresses this by lowering the FID score to 8.77.
Furthermore, while these SR baseline methods surpass the traditional fidelity-oriented methods~\cite{dong_compression_2015,wang_novel_2017,zhang_learning_2017,guo_toward_2019,zhang_residual_2018,xing_early_2020,zamir_multi-stage_2021} in most metrics, they fall short in PSNR, which does not aligned with perception as noted by \cite{blau_perception-distortion_2018}.
Nevertheless, our method also enhances the PSNR performance of these SR baseline methods by at least 0.6 dB.
This advancement is particularly meaningful to recent perception-driven quality enhancement endeavors, which predominantly employ SR baseline methods.
In summary, our method not only achieves state-of-the-art performance in enhancing the perceptual quality of compressed images, but also significantly improves the fidelity quality by perception-driven SR baseline methods.

\begin{figure*}
    \centering
    \includegraphics[width=\linewidth]{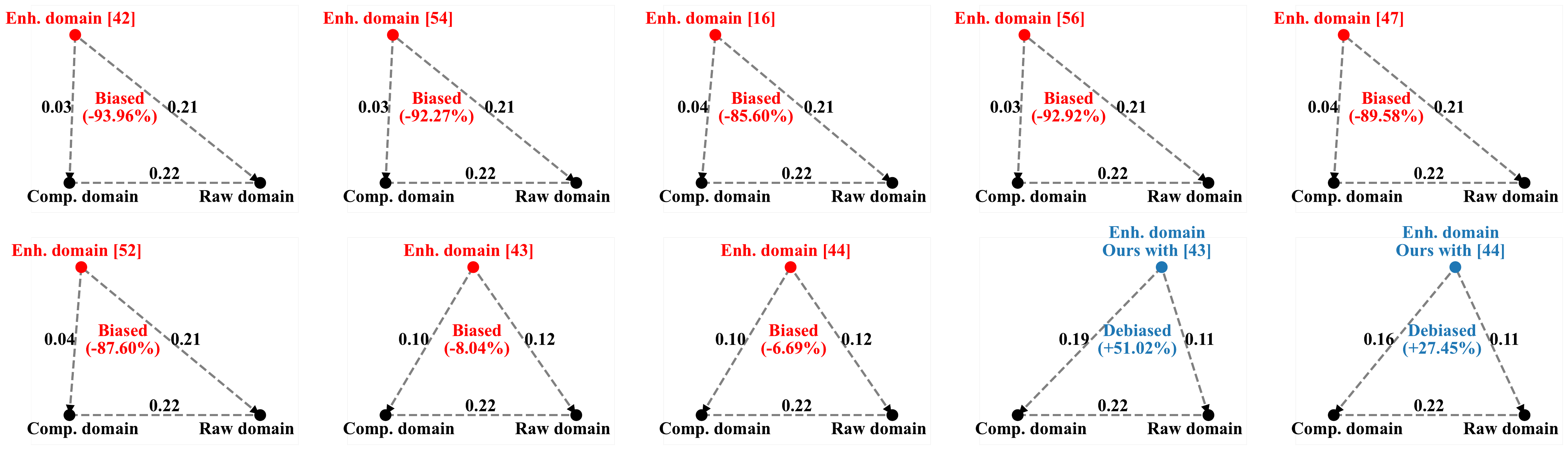}
    \caption{LPIPS scores between three domains on the evaluation set compressed by BPG with QP set to 37.
    The horizontal deviation of each vertex relative to the centroid of its base, calculated with float precision on original data, is also shown.}
    \label{fig:exp:bias_qp37_lpips}
    \vspace{\vspaces}
\end{figure*}

\begin{figure}
    \centering
    \includegraphics[width=\linewidth]{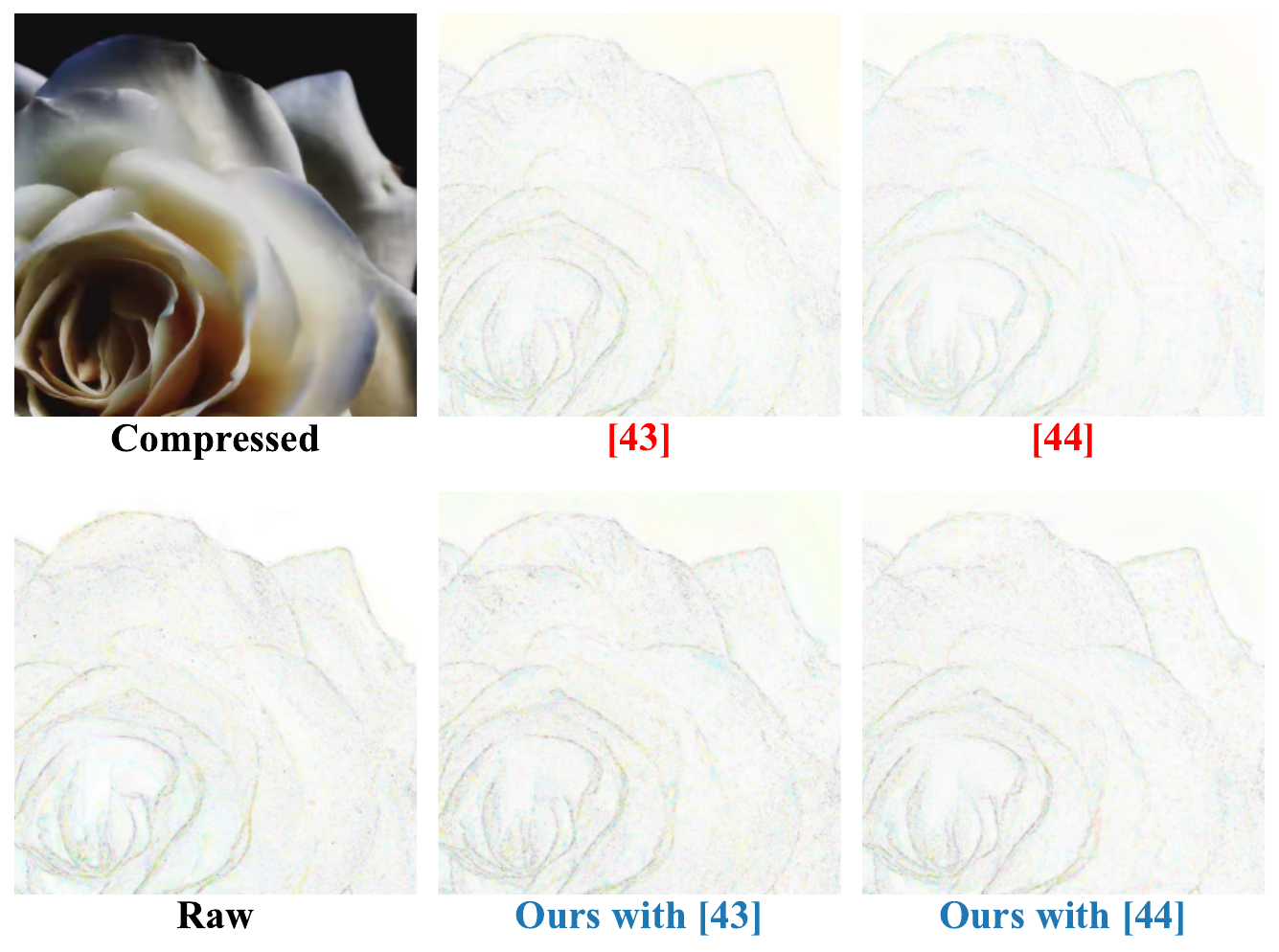}
    \caption{Visualization of residual to the compressed image.
    Additional examples are provided in the supplementary material.}
    \label{fig:exp:residual}
    \vspace{\vspacem}
\end{figure}

\textbf{Rate-distortion performance evaluation.}
In this section, we conduct a thorough evaluation of the rate-distortion performance of our and compared methods.
\cref{fig:exp:rd_curve} presents the rate-distortion curves for both BPG-compressed and JPEG-compressed images.
Notably, the curves corresponding to our method are consistently lower than those of the compared methods, highlighting the enhanced image quality by our method across codecs and bitrates.
Moreover, we quantify the rate-distortion performance through the Bjontegaard delta bit rate (BD-BR)~\cite{bjontegaard_calculation_2001} metric.
As illustrated in \cref{tab:exp:bdbr}, our method demonstrates a significant reduction in bitrate while maintaining the same perceptual quality compared to other methods, as evidenced by the minimal BD-BR values.
Even when considering PSNR, our method enhances the baseline performance by a minimum of 21.8\%.
In summary, our method significantly advances the state-of-the-art rate-perception performance, and also improves the rate-fidelity performance of perception-driven SR baseline methods.

\textbf{Enhancement bias mitigation.}
The foremost objective of our method is to mitigate the enhancement bias, thereby enhancing the quality of compressed images.
In this section, we examine the enhancement bias for both compared and our methods to validate the efficacy of our method in bias mitigation.
\cref{fig:exp:bias_qp37_fid,fig:exp:bias_qp37_lpips} present the similarity scores between three domains, as measured by FID and LPIPS respectively.
The results indicate that fidelity-oriented methods~\cite{wang_novel_2017,zhang_learning_2017,guo_toward_2019,zhang_residual_2018,xing_early_2020,zamir_multi-stage_2021} exhibit a pronounced enhancement bias towards the compression domain.
This is evident as their enhancement domain closely resembles the compression domain rather than the raw domain, a phenomenon observable in the left-leaning triangles.
In addition, employing perception-driven SR methods~\cite{wang_esrgan_2018,wang_real-esrgan_2021}, results in a modest reduction of the enhancement bias which remains noticeable.
Conversely, our proposed method substantially diminishes this bias, aligning the enhancement domain more closely with the raw domain.
\cref{fig:exp:residual} provides a visual demonstration, further illustrating the significant impact of our method in mitigating enhancement bias.
In conclusion, our proposed method proficiently addresses the prevalent enhancement bias found in existing methods.

\subsection{Ablation Study}

In our ablation study, we focus on two pivotal mechanisms of our proposed method: (1) the conditional discriminator, which aids in discerning the compression domain, and (2) the domain-divergence regularization, designed to more distinctly separate the enhancement domain from the compression domain.
As depicted in \cref{tab:exp:ablate}, the employment of either the conditional discriminator or the  domain-divergence regularization leads to improvements in both FID and PSNR metrics across various baseline methods, highlighting their efficacy in enhancing the quality of compressed images.
Moreover, our comprehensive method that integrates both mechanisms achieves the most significant advancements.

\begin{table}
    \centering
    \small
    \begin{tabular}{@{}l|cccc@{}}
        \toprule
        {Baseline} & {\cite{wang_esrgan_2018}} & {$\Delta$ to ori.} & {\cite{wang_real-esrgan_2021}} & {$\Delta$ to ori.} \\
        \midrule
        {Vanilla} & {8.84/29.0} & - & {10.6/29.8} & - \\
        {With (1)} & {8.05/29.4} & {-0.79/+0.38} & {9.21/30.0} & {-1.39/+0.24} \\
        {With (2)} & {7.67/29.2} & {-1.17/+0.22} & {9.70/30.4} & {-0.90/+0.59} \\
        {Ours \scriptsize{(Full)}} & \textbf{7.50/29.6} & \textbf{-1.34/+0.63} & \textbf{8.77/30.6} & \textbf{-1.83/+0.79} \\
        \bottomrule
    \end{tabular}
    \caption{FID/PSNR (dB) scores for enhanced BPG-compressed images with QP set to 37, using various configurations.}
    \label{tab:exp:ablate}
    \vspace{\vspacem}
\end{table}

\section{Conclusion}

In this paper, we systematically explored and addressed a critical issue prevalent in existing quality enhancement methods: the enhancement bias towards the compression domain, which detrimentally affects the quality of enhanced compressed images.
Our research introduced a simple yet effective method to mitigate this bias, thereby improving the quality of compressed images.
Our method utilizes a conditional discriminator for discerning the compression domain, enabling a clear distinction between the raw and enhancement domains.
This distinction is further reinforced by a domain-divergence regularization mechanism, effectively separating the enhancement domain from the compression domain.
Experimental results validated the effectiveness of our approach, showing significant improvement in image quality by more accurately aligning with the raw domain.
Our research not only achieves the state-of-the-art performance in quality enhancement, but also provides a novel perspective of evaluating the enhancement bias in future advancements of quality enhancement.

\noindent\textbf{Acknowledgements.} This work was supported by NSFC under Grants 62206011, 62250001, 62231002 and 62372024, Beijing Natural Science Foundation under Grant L223021, Academic Excellence Foundation of BUAA for PhD Students, and Alibaba Group through Alibaba Research Intern Program.

{
    \small
    \bibliographystyle{ieeenat_fullname}
    \bibliography{references}

\begin{thebibliography}{57}
\providecommand{\natexlab}[1]{#1}
\providecommand{\url}[1]{\texttt{#1}}
\expandafter\ifx\csname urlstyle\endcsname\relax
  \providecommand{\doi}[1]{doi: #1}\else
  \providecommand{\doi}{doi: \begingroup \urlstyle{rm}\Url}\fi

\bibitem[Agustsson and Timofte(2017)]{agustsson_ntire_2017}
Eirikur Agustsson and Radu Timofte.
\newblock {NTIRE} 2017 challenge on single image super-resolution: {Dataset} and study.
\newblock In \emph{2017 {IEEE} conference on computer vision and pattern recognition workshops, {CVPR} workshops 2017, honolulu, {HI}, {USA}, july 21-26, 2017}, pages 1122--1131. IEEE Computer Society, 2017.
\newblock tex.bibsource: dblp computer science bibliography, https://dblp.org tex.biburl: https://dblp.org/rec/conf/cvpr/AgustssonT17.bib tex.timestamp: Fri, 09 Apr 2021 18:48:31 +0200.

\bibitem[Arjovsky and Bottou(2017)]{arjovsky_towards_2017}
Martin Arjovsky and Leon Bottou.
\newblock Towards {Principled} {Methods} for {Training} {Generative} {Adversarial} {Networks}.
\newblock In \emph{International {Conference} on {Learning} {Representations}}, 2017.

\bibitem[Ba et~al.(2016)Ba, Kiros, and Hinton]{ba_layer_2016}
Jimmy~Lei Ba, Jamie~Ryan Kiros, and Geoffrey~E. Hinton.
\newblock Layer {Normalization}, 2016.
\newblock arXiv:1607.06450 [cs, stat].

\bibitem[Bellard(2018)]{bellard_better_2018}
Fabrice Bellard.
\newblock Better portable graphics ({BPG}), 2018.

\bibitem[Bjontegaard(2001)]{bjontegaard_calculation_2001}
Gisle Bjontegaard.
\newblock Calculation of average {PSNR} differences between {RD}-curves.
\newblock \emph{VCEG-M33}, 2001.

\bibitem[Blau and Michaeli(2018)]{blau_perception-distortion_2018}
Yochai Blau and Tomer Michaeli.
\newblock The perception-distortion tradeoff.
\newblock In \emph{2018 {IEEE}/{CVF} conference on computer vision and pattern recognition}. IEEE, 2018.

\bibitem[Blau et~al.(2018)Blau, Mechrez, Timofte, Michaeli, and Zelnik-Manor]{blau_2018_2018}
Yochai Blau, Roey Mechrez, Radu Timofte, Tomer Michaeli, and Lihi Zelnik-Manor.
\newblock The 2018 {PIRM} {Challenge} on {Perceptual} {Image} {Super}-{Resolution}.
\newblock In \emph{Proceedings of the {European} {Conference} on {Computer} {Vision} ({ECCV}) {Workshops}}, 2018.

\bibitem[Cai et~al.(2021)Cai, He, Qiao, and Dong]{cai_toward_2021}
Haoming Cai, Jingwen He, Yu Qiao, and Chao Dong.
\newblock Toward {Interactive} {Modulation} for {Photo}-{Realistic} {Image} {Restoration}.
\newblock In \emph{Proceedings of the {IEEE}/{CVF} {Conference} on {Computer} {Vision} and {Pattern} {Recognition} ({CVPR}) {Workshops}}, pages 294--303, 2021.

\bibitem[Chen et~al.(2023)Chen, Mo, Hou, Wu, Liao, Sun, Yan, and Lin]{chen_topiq_2023}
Chaofeng Chen, Jiadi Mo, Jingwen Hou, Haoning Wu, Liang Liao, Wenxiu Sun, Qiong Yan, and Weisi Lin.
\newblock {TOPIQ}: {A} {Top}-down {Approach} from {Semantics} to {Distortions} for {Image} {Quality} {Assessment}, 2023.
\newblock arXiv:2308.03060 [cs].

\bibitem[Ding et~al.(2020)Ding, Ma, Wang, and Simoncelli]{ding_image_2020}
Keyan Ding, Kede Ma, Shiqi Wang, and Eero~P. Simoncelli.
\newblock Image {Quality} {Assessment}: {Unifying} {Structure} and {Texture} {Similarity}.
\newblock \emph{IEEE Transactions on Pattern Analysis and Machine Intelligence}, pages 1--1, 2020.
\newblock arXiv:2004.07728 [cs].

\bibitem[Dong et~al.(2015)Dong, Deng, Loy, and Tang]{dong_compression_2015}
Chao Dong, Yubin Deng, Chen~Change Loy, and Xiaoou Tang.
\newblock Compression artifacts reduction by a deep convolutional network.
\newblock In \emph{2015 {IEEE} international conference on computer vision ({ICCV})}. IEEE, 2015.

\bibitem[Ghosh et~al.(2019)Ghosh, Hua, Mukherjee, and Robertson]{ghosh_iegan_2019}
Soumya Ghosh, Yang Hua, Sankha~Subhra Mukherjee, and Neil Robertson.
\newblock {IEGAN}: {Multi}-{Purpose} {Perceptual} {Quality} {Image} {Enhancement} {Using} {Generative} {Adversarial} {Network}.
\newblock In \emph{2019 {IEEE} {Winter} {Conference} on {Applications} of {Computer} {Vision} ({WACV})}, pages 11--20, 2019.
\newblock ISSN: 1550-5790.

\bibitem[Goodfellow et~al.(2014)Goodfellow, Pouget-Abadie, Mirza, Xu, Warde-Farley, Ozair, Courville, and Bengio]{goodfellow_generative_2014}
Ian Goodfellow, Jean Pouget-Abadie, Mehdi Mirza, Bing Xu, David Warde-Farley, Sherjil Ozair, Aaron Courville, and Yoshua Bengio.
\newblock Generative adversarial nets.
\newblock In \emph{Advances in neural information processing systems}. Curran Associates, Inc., 2014.

\bibitem[Guan et~al.(2021)Guan, Xing, Xu, Yang, Liu, and Wang]{guan_mfqe_2021}
Zhenyu Guan, Qunliang Xing, Mai Xu, Ren Yang, Tie Liu, and Zulin Wang.
\newblock {MFQE} 2.0: {A} {New} {Approach} for {Multi}-frame {Quality} {Enhancement} on {Compressed} {Video}.
\newblock \emph{IEEE Transactions on Pattern Analysis and Machine Intelligence}, 43\penalty0 (3):\penalty0 949--963, 2021.
\newblock arXiv:1902.09707 [cs].

\bibitem[Guo and Chao(2016)]{guo_building_2016}
Jun Guo and Hongyang Chao.
\newblock Building dual-domain representations for compression artifacts reduction.
\newblock In \emph{Computer vision – {ECCV} 2016}, pages 628--644. Springer International Publishing, 2016.

\bibitem[Guo et~al.(2019)Guo, Yan, Zhang, Zuo, and Zhang]{guo_toward_2019}
Shi Guo, Zifei Yan, Kai Zhang, Wangmeng Zuo, and Lei Zhang.
\newblock Toward convolutional blind denoising of real photographs.
\newblock In \emph{2019 {IEEE}/{CVF} conference on computer vision and pattern recognition ({CVPR})}. IEEE, 2019.

\bibitem[He et~al.(2016)He, Zhang, Ren, and Sun]{he_deep_2016}
Kaiming He, Xiangyu Zhang, Shaoqing Ren, and Jian Sun.
\newblock Deep residual learning for image recognition.
\newblock In \emph{2016 {IEEE} conference on computer vision and pattern recognition ({CVPR})}. IEEE, 2016.

\bibitem[Heusel et~al.(2017)Heusel, Ramsauer, Unterthiner, Nessler, and Hochreiter]{heusel_gans_2017}
Martin Heusel, Hubert Ramsauer, Thomas Unterthiner, Bernhard Nessler, and Sepp Hochreiter.
\newblock {GANs} trained by a two time-scale update rule converge to a local nash equilibrium.
\newblock In \emph{Proceedings of the 31st international conference on neural information processing systems}, pages 6629--6640, Red Hook, NY, USA, 2017. Curran Associates Inc.
\newblock Number of pages: 12 Place: Long Beach, California, USA.

\bibitem[Inc.(2020)]{inc_data_2020}
Domo Inc.
\newblock Data {Never} {Sleeps} 8.0: {How} much data is generated every minute?, 2020.

\bibitem[(ITU-T)(2017)]{itu-t_p10_2017}
International Telecommunication Union Communication Standardization~Sector (ITU-T).
\newblock P.10 : {Vocabulary} for performance, quality of service and quality of experience, 2017.

\bibitem[Jolicoeur-Martineau(2019)]{jolicoeur-martineau_relativistic_2019}
Alexia Jolicoeur-Martineau.
\newblock The relativistic discriminator: a key element missing from standard {GAN}.
\newblock In \emph{7th {International} {Conference} on {Learning} {Representations}, {ICLR} 2019, {New} {Orleans}, {LA}, {USA}, {May} 6-9, 2019}. OpenReview.net, 2019.

\bibitem[Ke et~al.(2021)Ke, Wang, Wang, Milanfar, and Yang]{ke_musiq_2021}
Junjie Ke, Qifei Wang, Yilin Wang, Peyman Milanfar, and Feng Yang.
\newblock {MUSIQ}: {Multi}-{Scale} {Image} {Quality} {Transformer}.
\newblock In \emph{Proceedings of the {IEEE}/{CVF} {International} {Conference} on {Computer} {Vision} ({ICCV})}, pages 5148--5157, 2021.

\bibitem[Kim et~al.(2020)Kim, Cho, Lee, Jeong, Choi, and Do]{kim_towards_2020}
Y. Kim, S. Cho, J. Lee, S. Jeong, J. Choi, and J. Do.
\newblock Towards the {Perceptual} {Quality} {Enhancement} of {Low} {Bit}-rate {Compressed} {Images}.
\newblock In \emph{2020 {IEEE}/{CVF} {Conference} on {Computer} {Vision} and {Pattern} {Recognition} {Workshops} ({CVPRW})}, pages 565--569, Los Alamitos, CA, USA, 2020. IEEE Computer Society.

\bibitem[Lao et~al.(2022)Lao, Gong, Shi, Yang, Wu, Wang, Xia, and Yang]{lao_attentions_2022}
Shanshan Lao, Yuan Gong, Shuwei Shi, Sidi Yang, Tianhe Wu, Jiahao Wang, Weihao Xia, and Yujiu Yang.
\newblock Attentions {Help} {CNNs} {See} {Better}: {Attention}-{Based} {Hybrid} {Image} {Quality} {Assessment} {Network}.
\newblock In \emph{Proceedings of the {IEEE}/{CVF} {Conference} on {Computer} {Vision} and {Pattern} {Recognition} ({CVPR}) {Workshops}}, pages 1140--1149, 2022.

\bibitem[Ledig et~al.(2017)Ledig, Theis, Huszar, Caballero, Cunningham, Acosta, Aitken, Tejani, Totz, Wang, and Shi]{ledig_photo-realistic_2017}
C. Ledig, L. Theis, F. Huszar, J. Caballero, A. Cunningham, A. Acosta, A. Aitken, A. Tejani, J. Totz, Z. Wang, and W. Shi.
\newblock Photo-{Realistic} {Single} {Image} {Super}-{Resolution} {Using} a {Generative} {Adversarial} {Network}.
\newblock In \emph{2017 {IEEE} {Conference} on {Computer} {Vision} and {Pattern} {Recognition} ({CVPR})}, pages 105--114, Los Alamitos, CA, USA, 2017. IEEE Computer Society.
\newblock ISSN: 1063-6919.

\bibitem[Li et~al.(2023)Li, Shi, and Yan]{li_iid-gan_2023}
Yang Li, Liangliang Shi, and Junchi Yan.
\newblock {IID}-{GAN}: an {IID} {Sampling} {Perspective} for {Regularizing} {Mode} {Collapse}.
\newblock In \emph{Proceedings of the {Thirty}-{Second} {International} {Joint} {Conference} on {Artificial} {Intelligence}}, pages 3929--3938, Macau, SAR China, 2023. International Joint Conferences on Artificial Intelligence Organization.

\bibitem[Marcellin et~al.(2000)Marcellin, Gormish, Bilgin, and Boliek]{marcellin_overview_2000}
M.W. Marcellin, M.J. Gormish, A. Bilgin, and M.P. Boliek.
\newblock An overview of {JPEG}-2000.
\newblock In \emph{Proceedings {DCC} 2000. {Data} compression conference}. IEEE Comput. Soc, 2000.

\bibitem[Mentzer et~al.(2020)Mentzer, Toderici, Tschannen, and Agustsson]{mentzer_high-fidelity_2020}
Fabian Mentzer, George Toderici, Michael Tschannen, and Eirikur Agustsson.
\newblock High-{Fidelity} {Generative} {Image} {Compression}, 2020.
\newblock arXiv:2006.09965 [cs, eess].

\bibitem[Mirza and Osindero(2014)]{mirza_conditional_2014}
Mehdi Mirza and Simon Osindero.
\newblock Conditional {Generative} {Adversarial} {Nets}, 2014.
\newblock arXiv:1411.1784 [cs, stat].

\bibitem[Mittal et~al.(2013)Mittal, Soundararajan, and Bovik]{mittal_making_2013}
Anish Mittal, Rajiv Soundararajan, and Alan~C. Bovik.
\newblock Making a “{Completely} {Blind}” {Image} {Quality} {Analyzer}.
\newblock \emph{IEEE Signal Processing Letters}, 20\penalty0 (3):\penalty0 209--212, 2013.
\newblock Conference Name: IEEE Signal Processing Letters.

\bibitem[Miyato et~al.(2018)Miyato, Kataoka, Koyama, and Yoshida]{miyato_spectral_2018}
Takeru Miyato, Toshiki Kataoka, Masanori Koyama, and Yuichi Yoshida.
\newblock Spectral {Normalization} for {Generative} {Adversarial} {Networks}.
\newblock In \emph{International {Conference} on {Learning} {Representations}}, 2018.

\bibitem[Schonfeld et~al.(2020)Schonfeld, Schiele, and Khoreva]{schonfeld_u-net_2020}
E. Schonfeld, B. Schiele, and A. Khoreva.
\newblock A {U}-{Net} {Based} {Discriminator} for {Generative} {Adversarial} {Networks}.
\newblock In \emph{2020 {IEEE}/{CVF} {Conference} on {Computer} {Vision} and {Pattern} {Recognition} ({CVPR})}, pages 8204--8213, Los Alamitos, CA, USA, 2020. IEEE Computer Society.

\bibitem[Seshadrinathan et~al.(2010)Seshadrinathan, Soundararajan, Bovik, and Cormack]{seshadrinathan_study_2010}
Kalpana Seshadrinathan, Rajiv Soundararajan, Alan~Conrad Bovik, and Lawrence~K. Cormack.
\newblock Study of subjective and objective quality assessment of video.
\newblock \emph{IEEE Transactions on Image Processing}, 19\penalty0 (6):\penalty0 1427--1441, 2010.
\newblock Publisher: Institute of Electrical and Electronics Engineers (IEEE).

\bibitem[Shen and Kuo(1998)]{shen_review_1998}
Mei-Yin Shen and C.~C.~Jay Kuo.
\newblock Review of {Postprocessing} {Techniques} for {Compression} {Artifact} {Removal}.
\newblock \emph{Journal of Visual Communication and Image Representation}, 9\penalty0 (1):\penalty0 2--14, 1998.

\bibitem[Simonyan and Zisserman(2015)]{simonyan_very_2015}
Karen Simonyan and Andrew Zisserman.
\newblock Very deep convolutional networks for large-scale image recognition.
\newblock In \emph{3rd international conference on learning representations, {ICLR} 2015, san diego, {CA}, {USA}, may 7-9, 2015, conference track proceedings}, 2015.
\newblock tex.bibsource: dblp computer science bibliography, https://dblp.org tex.biburl: https://dblp.org/rec/journals/corr/SimonyanZ14a.bib tex.timestamp: Wed, 17 Jul 2019 10:40:54 +0200.

\bibitem[Su et~al.(2020)Su, Yan, Zhu, Zhang, Ge, Sun, and Zhang]{su_blindly_2020}
Shaolin Su, Qingsen Yan, Yu Zhu, Cheng Zhang, Xin Ge, Jinqiu Sun, and Yanning Zhang.
\newblock Blindly {Assess} {Image} {Quality} in the {Wild} {Guided} by a {Self}-{Adaptive} {Hyper} {Network}.
\newblock In \emph{2020 {IEEE}/{CVF} {Conference} on {Computer} {Vision} and {Pattern} {Recognition} ({CVPR})}, pages 3664--3673, 2020.
\newblock ISSN: 2575-7075.

\bibitem[Sullivan et~al.(2012)Sullivan, Ohm, Han, and Wiegand]{sullivan_overview_2012}
Gary~J. Sullivan, Jens-Rainer Ohm, Woo-Jin Han, and Thomas Wiegand.
\newblock Overview of the high efficiency video coding ({HEVC}) standard.
\newblock \emph{IEEE Transactions on Circuits and Systems for Video Technology}, 22\penalty0 (12):\penalty0 1649--1668, 2012.
\newblock Publisher: Institute of Electrical and Electronics Engineers (IEEE).

\bibitem[Thulasiraman and Swamy(1992)]{thulasiraman_graphs_1992}
K. Thulasiraman and M.N.S. Swamy.
\newblock \emph{Graphs: {Theory} and {Algorithms}}.
\newblock Wiley, 1992.

\bibitem[Timofte et~al.(2017)Timofte, Agustsson, Gool, Yang, Zhang, Lim, Son, Kim, Nah, Lee, Wang, Tian, Yu, Zhang, Wu, Dong, Lin, Qiao, Loy, Bae, Yoo, Han, Ye, Choi, Kim, Fan, Yu, Han, Liu, Yu, Wang, Shi, Wang, Huang, Chen, Zhang, Zuo, Tang, Luo, Li, Fu, Cao, Heng, Bui, Le, Duan, Tao, Wang, Lin, Pang, Xu, Zhao, Xu, Pan, Sun, Zhang, Song, Dai, Qin, Huynh, Guo, Mousavi, Vu, Monga, Cruz, Egiazarian, Katkovnik, Mehta, Jain, Agarwalla, Praveen, Zhou, Wen, Zhu, Xia, Wang, and Guo]{timofte_ntire_2017}
Radu Timofte, Eirikur Agustsson, Luc~Van Gool, Ming-Hsuan Yang, Lei Zhang, Bee Lim, Sanghyun Son, Heewon Kim, Seungjun Nah, Kyoung~Mu Lee, Xintao Wang, Yapeng Tian, Ke Yu, Yulun Zhang, Shixiang Wu, Chao Dong, Liang Lin, Yu Qiao, Chen~Change Loy, Woong Bae, Jae~Jun Yoo, Yoseob Han, Jong~Chul Ye, Jae-Seok Choi, Munchurl Kim, Yuchen Fan, Jiahui Yu, Wei Han, Ding Liu, Haichao Yu, Zhangyang Wang, Honghui Shi, Xinchao Wang, Thomas~S. Huang, Yunjin Chen, Kai Zhang, Wangmeng Zuo, Zhimin Tang, Linkai Luo, Shaohui Li, Min Fu, Lei Cao, Wen Heng, Giang Bui, Truc Le, Ye Duan, Dacheng Tao, Ruxin Wang, Xu Lin, Jianxin Pang, Jinchang Xu, Yu Zhao, Xiangyu Xu, Jin-shan Pan, Deqing Sun, Yujin Zhang, Xibin Song, Yuchao Dai, Xueying Qin, Xuan-Phung Huynh, Tiantong Guo, Hojjat~Seyed Mousavi, Tiep~Huu Vu, Vishal Monga, Cristóvão Cruz, Karen~O. Egiazarian, Vladimir Katkovnik, Rakesh Mehta, Arnav~Kumar Jain, Abhinav Agarwalla, Ch~V.~Sai Praveen, Ruofan Zhou, Hongdiao Wen, Che Zhu, Zhiqiang Xia, Zhengtao Wang, and Qi Guo.
\newblock {NTIRE} 2017 challenge on single image super-resolution: {Methods} and results.
\newblock In \emph{2017 {IEEE} conference on computer vision and pattern recognition workshops, {CVPR} workshops 2017, honolulu, {HI}, {USA}, july 21-26, 2017}, pages 1110--1121. IEEE Computer Society, 2017.
\newblock tex.bibsource: dblp computer science bibliography, https://dblp.org tex.biburl: https://dblp.org/rec/conf/cvpr/TimofteAG0ZLSKN17.bib tex.timestamp: Thu, 21 Apr 2022 09:15:18 +0200.

\bibitem[Wallace(1992)]{wallace_jpeg_1992}
G.K. Wallace.
\newblock The {JPEG} still picture compression standard.
\newblock \emph{IEEE Transactions on Consumer Electronics}, 38\penalty0 (1):\penalty0 xviii--xxxiv, 1992.
\newblock Publisher: Institute of Electrical and Electronics Engineers (IEEE).

\bibitem[Wang et~al.(2023)Wang, Chan, and Loy]{wang_exploring_2023}
Jianyi Wang, Kelvin C.~K. Chan, and Chen~Change Loy.
\newblock Exploring {CLIP} for {Assessing} the {Look} and {Feel} of {Images}.
\newblock \emph{Proceedings of the AAAI Conference on Artificial Intelligence}, 37\penalty0 (2):\penalty0 2555--2563, 2023.
\newblock Number: 2.

\bibitem[Wang et~al.(2017)Wang, Chen, and Chao]{wang_novel_2017}
Tingting Wang, Mingjin Chen, and Hongyang Chao.
\newblock A novel deep learning-based method of improving coding efficiency from the decoder-end for {HEVC}.
\newblock In \emph{2017 data compression conference ({DCC})}. IEEE, 2017.

\bibitem[Wang et~al.(2018)Wang, Yu, Wu, Gu, Liu, Dong, Qiao, and Change~Loy]{wang_esrgan_2018}
Xintao Wang, Ke Yu, Shixiang Wu, Jinjin Gu, Yihao Liu, Chao Dong, Yu Qiao, and Chen Change~Loy.
\newblock {ESRGAN}: {Enhanced} {Super}-{Resolution} {Generative} {Adversarial} {Networks}.
\newblock In \emph{Proceedings of the {European} {Conference} on {Computer} {Vision} ({ECCV}) {Workshops}}, 2018.

\bibitem[Wang et~al.(2021)Wang, Xie, Dong, and Shan]{wang_real-esrgan_2021}
Xintao Wang, Liangbin Xie, Chao Dong, and Ying Shan.
\newblock Real-{ESRGAN}: {Training} {Real}-{World} {Blind} {Super}-{Resolution} {With} {Pure} {Synthetic} {Data}.
\newblock In \emph{Proceedings of the {IEEE}/{CVF} {International} {Conference} on {Computer} {Vision} ({ICCV}) {Workshops}}, pages 1905--1914, 2021.

\bibitem[Wang et~al.(2016)Wang, Liu, Chang, Ling, Yang, and Huang]{wang_d3_2016}
Zhangyang Wang, Ding Liu, Shiyu Chang, Qing Ling, Yingzhen Yang, and Thomas~S. Huang.
\newblock D3: {Deep} dual-domain based fast restoration of {JPEG}-{Compressed} images.
\newblock In \emph{2016 {IEEE} conference on computer vision and pattern recognition ({CVPR})}. IEEE, 2016.

\bibitem[{Wikipedia contributors}(2023)]{wikipedia_contributors_triangle_2023}
{Wikipedia contributors}.
\newblock Triangle inequality — {Wikipedia}, {The} {Free} {Encyclopedia}, 2023.

\bibitem[Xing et~al.(2020)Xing, Xu, Li, and Guan]{xing_early_2020}
Qunliang Xing, Mai Xu, Tianyi Li, and Zhenyu Guan.
\newblock Early exit or not: {Resource}-efficient blind quality enhancement for compressed images.
\newblock In \emph{Computer vision - {ECCV} 2020 - 16th european conference, glasgow, {UK}, august 23-28, 2020, proceedings, part {XVI}}, pages 275--292. Springer, 2020.
\newblock tex.bibsource: dblp computer science bibliography, https://dblp.org tex.biburl: https://dblp.org/rec/conf/eccv/XingXLG20.bib tex.timestamp: Thu, 17 Feb 2022 16:43:16 +0100.

\bibitem[Xing et~al.(2023)Xing, Xu, Deng, and Guo]{xing_daqe_2023}
Qunliang Xing, Mai Xu, Xin Deng, and Yichen Guo.
\newblock {DAQE}: {Enhancing} the {Quality} of {Compressed} {Images} by {Exploiting} the {Inherent} {Characteristic} of {Defocus}.
\newblock \emph{IEEE Transactions on Pattern Analysis and Machine Intelligence}, pages 1--17, 2023.
\newblock Conference Name: IEEE Transactions on Pattern Analysis and Machine Intelligence.

\bibitem[Yang et~al.(2018)Yang, Xu, Wang, and Li]{yang_multi-frame_2018}
Ren Yang, Mai Xu, Zulin Wang, and Tianyi Li.
\newblock Multi-frame quality enhancement for compressed video.
\newblock In \emph{2018 {IEEE}/{CVF} conference on computer vision and pattern recognition}. IEEE, 2018.

\bibitem[Yang et~al.(2022)Yang, Timofte, and Van~Gool]{yang_perceptual_2022}
Ren Yang, Radu Timofte, and Luc Van~Gool.
\newblock Perceptual {Learned} {Video} {Compression} with {Recurrent} {Conditional} {GAN}, 2022.
\newblock arXiv:2109.03082 [cs, eess].

\bibitem[Yu et~al.(2019)Yu, Qu, and Hong]{yu_underwater-gan_2019}
Xiaoli Yu, Yanyun Qu, and Ming Hong.
\newblock Underwater-{GAN}: {Underwater} {Image} {Restoration} via {Conditional} {Generative} {Adversarial} {Network}.
\newblock In \emph{Pattern {Recognition} and {Information} {Forensics}}, pages 66--75, Cham, 2019. Springer International Publishing.

\bibitem[Zamir et~al.(2021)Zamir, Arora, Khan, Hayat, Khan, Yang, and Shao]{zamir_multi-stage_2021}
Syed~Waqas Zamir, Aditya Arora, Salman~H. Khan, Munawar Hayat, Fahad~Shahbaz Khan, Ming-Hsuan Yang, and Ling Shao.
\newblock Multi-stage progressive image restoration.
\newblock In \emph{{IEEE} conference on computer vision and pattern recognition, {CVPR} 2021, virtual, june 19-25, 2021}, pages 14821--14831. Computer Vision Foundation / IEEE, 2021.
\newblock tex.bibsource: dblp computer science bibliography, https://dblp.org tex.biburl: https://dblp.org/rec/conf/cvpr/ZamirA0HK0021.bib tex.timestamp: Mon, 30 Aug 2021 17:00:27 +0200.

\bibitem[Zhang et~al.(2020)Zhang, Sindagi, and Patel]{zhang_image_2020}
He Zhang, Vishwanath Sindagi, and Vishal~M. Patel.
\newblock Image {De}-{Raining} {Using} a {Conditional} {Generative} {Adversarial} {Network}.
\newblock \emph{IEEE Transactions on Circuits and Systems for Video Technology}, 30\penalty0 (11):\penalty0 3943--3956, 2020.
\newblock Conference Name: IEEE Transactions on Circuits and Systems for Video Technology.

\bibitem[Zhang et~al.(2017)Zhang, Zuo, Gu, and Zhang]{zhang_learning_2017}
K. Zhang, W. Zuo, S. Gu, and L. Zhang.
\newblock Learning {Deep} {CNN} {Denoiser} {Prior} for {Image} {Restoration}.
\newblock In \emph{2017 {IEEE} {Conference} on {Computer} {Vision} and {Pattern} {Recognition} ({CVPR})}, pages 2808--2817, Los Alamitos, CA, USA, 2017. IEEE Computer Society.
\newblock ISSN: 1063-6919.

\bibitem[Zhang et~al.(2018{\natexlab{a}})Zhang, Isola, Efros, Shechtman, and Wang]{zhang_unreasonable_2018}
Richard Zhang, Phillip Isola, Alexei~A. Efros, Eli Shechtman, and Oliver Wang.
\newblock The unreasonable effectiveness of deep features as a perceptual metric.
\newblock In \emph{2018 {IEEE}/{CVF} conference on computer vision and pattern recognition}. IEEE, 2018{\natexlab{a}}.

\bibitem[Zhang et~al.(2018{\natexlab{b}})Zhang, Tian, Kong, Zhong, and Fu]{zhang_residual_2018}
Yulun Zhang, Yapeng Tian, Yu Kong, Bineng Zhong, and Yun Fu.
\newblock Residual {Dense} {Network} for {Image} {Super}-{Resolution}.
\newblock In \emph{Proceedings of the {IEEE} {Conference} on {Computer} {Vision} and {Pattern} {Recognition} ({CVPR})}, 2018{\natexlab{b}}.

\bibitem[Zheng et~al.(2022)Zheng, Xing, Qiao, Xu, Jiang, Liu, and Chen]{zheng_progressive_2022}
Meisong Zheng, Qunliang Xing, Minglang Qiao, Mai Xu, Lai Jiang, Huaida Liu, and Ying Chen.
\newblock Progressive {Training} of {A} {Two}-{Stage} {Framework} for {Video} {Restoration}.
\newblock In \emph{2022 {IEEE}/{CVF} {Conference} on {Computer} {Vision} and {Pattern} {Recognition} {Workshops} ({CVPRW})}, pages 1023--1030, New Orleans, LA, USA, 2022. IEEE.

\end{thebibliography}
}

\end{document}